\documentclass{article}

\usepackage[preprint]{neurips_2026}

\usepackage[utf8]{inputenc}
\usepackage[T1]{fontenc}
\usepackage{hyperref}
\usepackage{url}
\usepackage{booktabs}
\usepackage{amsfonts}
\usepackage{amsmath}
\usepackage{nicefrac}
\usepackage{microtype}
\usepackage{graphicx}
\usepackage{xcolor}
\usepackage{tikz}
\usetikzlibrary{arrows.meta, positioning, calc}

\title{Input-Adaptive Gating of a Dehazing Front-End for\\ On-Device Perception in Smoke-Obscured Environments}

\author{%
  Seongjun Kang \qquad Ishaan Garg \qquad Vishnu Bharadwaj \\
  Monta Vista High School \\
  \texttt{\{seongjunkang1, ishaangarg323, vishnubharadwaj83\}@gmail.com}
}

\begin{document}

\maketitle

\begin{abstract}
Two-stage vision pipelines often place an enhancement network before a task network, on the assumption that a cleaner input produces a better output. We evaluate this in a firefighter assistance pipeline, where a dehazer precedes an edge detector that renders smoke-filled rooms as structural outlines. Both were designed for a Raspberry Pi 4, at 355K and 23K parameters, and quantized to UINT8 via TensorFlow Lite. The float dehazer reaches 18.60 dB peak signal-to-noise ratio (PSNR) on held-out real smoke against 13.60 dB unprocessed and 17.08 dB for an AOD-Net trained on the same data, and the edge detector reaches an F-measure at optimal dataset scale (ODS) of 0.738, outperforming an optimized Canny's result of 0.692. Dehazing improves edge extraction under dense smoke but degrades it on clear and lightly hazed frames, where the dehazer discards more detail than the haze obscures. We therefore run the dehazer only when a dark channel haze estimate exceeds a threshold, a 10.1 ms test that lets the pipeline save 469.6 ms on the dehazing stage. Averaged over four haze levels, gating is more accurate than either fixed decision, at 0.675 mean ODS against 0.664 for always dehazing and 0.630 for never dehazing. It reduces the mean per-frame time on the Raspberry Pi from 569 ms to 321 ms, and on clear frames increases the frame rate fivefold, from 1.8 to 9 frames per second. \end{abstract}

\section{Introduction}
\label{sec:intro}

Approximately 100 firefighters die on duty in the United States each year, with limited visibility in smoke-filled environments being a leading factor~\citep{usfa2022fatalities}. Equipment carried into a burning structure has known limitations in dense smoke: thermal imaging cameras are expensive and lose accuracy in cluttered scenes, flashlights backscatter off smoke particles, and night vision devices bloom near flames.

A fire ground is also a difficult environment for cloud-dependent connection. Buildings attenuate radio signals, infrastructure may be damaged or powered down, and a navigation aid cannot handle round-trip latency or dropped connections. For this reason, we built a two-stage computer vision pipeline that runs offline on a Raspberry Pi 4 Model B: a dehazing neural network removes smoke from the camera frame, and an edge detection model converts the result into a high-contrast outline of walls, doorways, and obstacles.

This design follows a common pattern in which an enhancement network precedes a task network, assuming an improvement downstream. This assumption is rarely tested because on a workstation an unnecessary enhancement pass is cheap. However, the same does not apply to embedded hardware as the dehazer accounts for 83\% of our pipeline's measured per-frame latency.

We measured how dehazing affects edge accuracy across a range of haze densities, and conditioned the dehazing stage on that measurement. The main contributions of this work are:

\begin{itemize}\setlength{\itemsep}{2pt}\setlength{\parskip}{0pt}
  \item A test of the assumption that a cleaner input gives a better output. It holds under dense smoke, where dehazing improves edge accuracy, but fails on clear and lightly hazed frames, where dehazing degrades it.
  \item A gate that runs the 470 ms dehazer only when a 10 ms dark channel estimate of haze density exceeds a fitted threshold. It reaches higher mean edge accuracy than always or never dehazing, and cuts per-frame cost by 43.5\%.
  \item A two-stage pipeline that runs entirely on a Raspberry Pi 4, at 355K and 23K parameters, with both networks exported to full-integer TensorFlow Lite.
\end{itemize}

\section{Related work}
\label{sec:related}

Existing dehazing methods include the dark channel prior~\citep{he2011dcp}, which estimates haze from the darkest pixel in each local patch, and small convolutional networks such as DehazeNet~\citep{cai2016dehazenet} and AOD-Net~\citep{li2017aodnet}. All three are small enough for embedded deployment, and we use
the two learned models as baselines. For edge detection, Canny~\citep{canny1986computational}
is the standard computer vision algorithm, while HED~\citep{xie2015holistically} and
DexiNed~\citep{poma2020dexined,soria2023dexined} are convolutional models that produce cleaner
contours. Both are built for workstation GPUs and exceed our parameter budget, so we use Canny as
our quantitative baseline. To keep our own networks small, we use depthwise separable
convolutions~\citep{howard2017mobilenets,chollet2017xception} and integer
quantization~\citep{jacob2018quantization}, standard tools for reducing vision models to mobile scale.

\section{Method}
\label{sec:system}

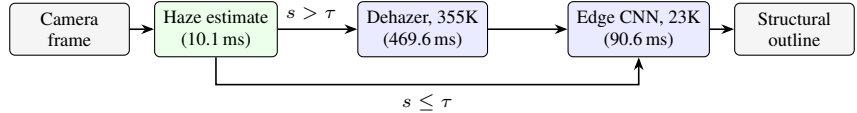
\begin{figure}[h]
  \centering
  \begin{tikzpicture}[
    node distance=0.32cm,
    box/.style={draw, rounded corners=2pt, minimum height=0.64cm, minimum width=1.58cm, align=center, font=\scriptsize},
    model/.style={box, fill=blue!8},
    data/.style={box, fill=gray!8},
    gatebox/.style={box, fill=green!8},
    arr/.style={-{Stealth[length=1.7mm]}, semithick},
    lbl/.style={font=\scriptsize}
  ]
    \node[data] (cam) {Camera\\frame};
    \node[gatebox, right=of cam] (gate) {Haze estimate\\(10.1\,ms)};
    \node[model, right=1.05cm of gate] (dehaze) {Dehazer, 355K\\(469.6\,ms)};
    \node[model, right=1.05cm of dehaze] (edge) {Edge CNN, 23K\\(90.6\,ms)};
    \node[data, right=of edge] (out) {Structural\\outline};
    \draw[arr] (cam) -- (gate);
    \draw[arr] (gate) -- node[lbl, above, pos=0.42] {$s>\tau$} (dehaze);
    \draw[arr] (dehaze) -- (edge);
    \draw[arr] (edge) -- (out);
    \draw[arr] (gate.south) -- ++(0,-0.36) -- node[lbl, below, pos=0.5] {$s\leq\tau$} ($(edge.south)+(0,-0.36)$) -- (edge.south);
  \end{tikzpicture}
  \caption{The gated pipeline. The dehazing stage runs only when the dark channel haze estimate $s$ exceeds the threshold $\tau$. Latencies are per-frame medians on a Raspberry Pi 4 Model B (100 iterations per network, 200 for the gate).}
  \label{fig:pipeline}
\end{figure}

\begin{figure}[t]
  \centering
  \includegraphics[width=0.82\linewidth]{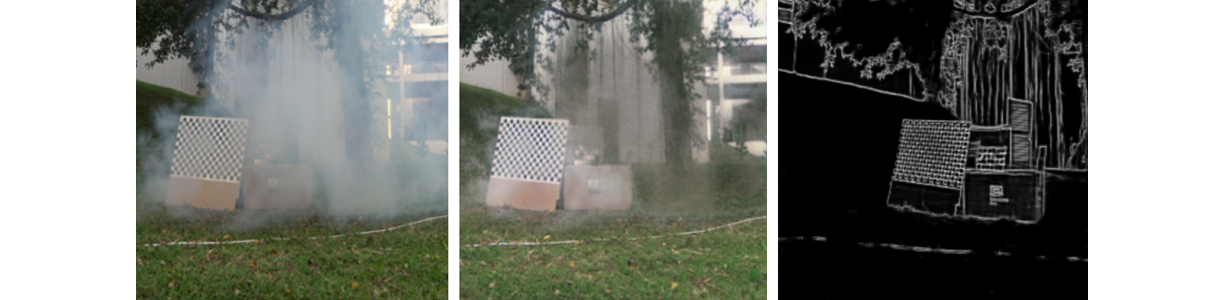}
  \caption{The pipeline on a real smoke capture: smoky input, dehazed frame, edge map.}
  \label{fig:qualitative}
\end{figure}

The pipeline has two networks and a gate (Figure~\ref{fig:pipeline}). A dehazer removes smoke from the input, and an edge detector renders the result as a high-contrast outline (Figure~\ref{fig:qualitative}). The gate decides whether the dehazer runs at all, or whether the frame goes straight to the edge detector.

\paragraph{Smoke removal network.} The dehazing model includes convolutional encoders and decoders with 355{,}043 parameters trained to regress the clean image from the hazy input. The encoder downsamples twice and then applies a dilated convolution that widens the receptive field without reducing the spatial size. The decoder restores resolution with two transposed convolutions and concatenates skip features from the matching encoder stages. Appendix~\ref{app:arch} gives the layer specification.

\paragraph{Edge detection network.} The edge network is a U-Net-style~\citep{ronneberger2015unet} model with 23{,}228 parameters, trained with binary cross-entropy loss against human-annotated edge maps. The two encoder blocks of depthwise separable convolutions~\citep{chollet2017xception} reduce resolution by $4\times$, and the decoder restores it with nearest-neighbor upsampling and skip concatenation (Appendix~\ref{app:arch}).
To deploy the model on a Raspberry Pi, we use separable convolutions to reduce parameter count and compute cost, and nearest-neighbor upsampling to avoid a TensorFlow Lite conversion failure on the target runtime (Appendix~\ref{app:opset}).

\paragraph{Embedded deployment.} Both networks' weights, activations, and inference inputs are quantized to uint8 with TensorFlow Lite~\citep{abadi2015tensorflow,jacob2018quantization}. The quantized edge model is 56 KB and the dehazer 373 KB; the re-export that loads on the device is 53 KB (Appendix~\ref{app:opset}). The platform is a Raspberry Pi 4 Model B (quad-core Cortex-A72) on 32-bit Raspberry Pi OS with Python 3.7, reading a USB webcam through OpenCV. Frames are resized, run through the interpreter, normalized, and displayed.

\subsection{Haze-conditioned gate}
\label{sec:gate}

In atmospheric scattering models, the dark channel of an image, a minimum pixel value within a $15\times15$ patch, is near zero in clear scenes and rises towards atmospheric light as transmission falls~\citep{he2011dcp}. To estimate the haze density, we calculate the dark channel prior mean $s$ and only run the dehazer when $s$ exceeds a threshold $\tau$. We test candidate thresholds on 60 BIPEDv2 training images at the four haze levels of Section~\ref{sec:gateresults}, keeping $\tau$ with the greatest total gain in the edge F-measure: $\tau=0.585$. Computing $s$ requires 0.002 giga multiply-accumulate operations (GMAC) and takes 10.1 ms with an OpenCV~\citep{bradski2000opencv} morphological filter, costing 2\% of the dehazer's latency

\section{Experimental setup}
\label{sec:setup}

\paragraph{Datasets.}We first trained the dehazer on 1,399 synthetically hazed images from RESIDE~\citep{li2019reside}. An experienced firefighter reviewed early outputs and objected that synthetic haze does not resemble real fire conditions, so we retrained on SMOKE~\citep{jin2022structure} (110 training and 12 test real smoke pairs) and DENSE-HAZE~\citep{ancuti2019dense,ancuti2019ntire} (55 pairs), trading dataset size for realism. The edge model uses BIPEDv2~\citep{soria2023dexined} (200 training and 50 test images) with
hand-labeled edge maps as ground truth.

\paragraph{Training.} Images were resized to $256\times256$ and normalized to $[0,1]$. Both training
sets were augmented with random flips and rotations in $[-180^\circ,180^\circ]$, applied identically
to each image and its label, adding seven copies per dehazing pair and five per BIPEDv2 image. Our
models and the baselines used Adam~\citep{kingma2015adam} at $10^{-3}$, batch size 16, early
stopping with patience 5, and a 20\% validation split, over at most 50 epochs for the dehazer and 30
for the edge network. Our DehazeNet reimplementation replaces the published clipped transmission head with a scaled sigmoid and trains at $10^{-4}$. With the clipped head at $10^{-3}$, the transmission map saturates at its upper bound and the network converges to the identity.

\paragraph{Metrics.} For dehazing, we report peak signal-to-noise ratio (PSNR), which compares similarity in pixel values, and structural similarity index measure (SSIM)~\citep{wang2004ssim}, which tracks perceived quality more closely. For edge detection we report F-measure at optimal dataset and image scale (ODS and OIS)~\citep{arbelaez2011contour,soria2023dexined}: predictions and ground truth are thinned to single-pixel width and matched within 2 px by bipartite assignment, so each annotated pixel matches at most one predicted pixel. Canny~\citep{canny1986computational} is swept over its upper hysteresis threshold, with the lower fixed at half, and we report its best operating point. \paragraph{Availability.} Code and the Raspberry Pi deployment kit will be released. Appendix~\ref{app:release} covers dataset licences, training compute, and broader impact.

\section{Results}
\label{sec:results}

\paragraph{Smoke removal.} Our model reaches 18.60 dB on the 12 held-out SMOKE pairs (Table~\ref{tab:main}, left), the best result in the comparison on both metrics and 5.0 dB above the untouched input; AOD-Net follows at 17.08 dB. The dark channel prior falls below the input: its assumption that every local patch contains a dark pixel does not hold when the air is thick and bright, so it over-estimates the haze and darkens the image about twice as much as it should. That assumption was built for outdoor haze, not for smoke, so we learn the correction from data instead.

\paragraph{Edge detection.} The deployed 23K parameter model reaches ODS 0.738 on the BIPEDv2 test split, against 0.692 for the best swept Canny operating point (Table~\ref{tab:main}, right). A 1.95M-parameter version reaches 0.786, so the deployed model is within 0.048 ODS of it while being 84 times smaller. Full-integer conversion moves the edge network to ODS 0.740, a change within noise, while the same quantization methods cost the dehazer 1.8 dB PSNR (Appendix~\ref{app:quant}).

\begin{table}[t]
  \caption{\textbf{Left:} smoke removal on the 12 held-out SMOKE test pairs, against the dark channel prior~\citep{he2011dcp}, DehazeNet~\citep{cai2016dehazenet} and AOD-Net~\citep{li2017aodnet}; we reimplemented both learned baselines and trained them on the same 165 pairs and augmentation as our model (Section~\ref{sec:setup}). \textbf{Right:} edge detection on the BIPEDv2 test split (50 images) using one-to-one pixel matching (Section~\ref{sec:setup}), with Canny~\citep{canny1986computational} swept over its hysteresis threshold.}
  \label{tab:main}
  \centering
  \footnotesize
  \begin{minipage}[t]{0.515\linewidth}
    \centering
    \scriptsize
    \begin{tabular}{lrcc}
      \toprule
      Method & Params & PSNR & SSIM \\
      \midrule
      Hazy input (no dehazing)   & --  & 13.60 & 0.507 \\
      Dark channel prior         & --  & 13.04 & 0.515 \\
      DehazeNet     & 8.3K  & 15.85 & 0.534 \\
      AOD-Net                    & 1.8K  & 17.08 & 0.580 \\
      Ours (float32)             & 355K  & \textbf{18.60} & \textbf{0.616} \\
      Ours (uint8, 373 KB)       & 355K  & 16.80 & 0.462 \\
      \bottomrule
    \end{tabular}
  \end{minipage}\hfill
  \begin{minipage}[t]{0.475\linewidth}
    \centering
    \scriptsize
    \begin{tabular}{lrcc}
      \toprule
      Method & Params & ODS & OIS \\
      \midrule
      Canny (best of sweep)          & --    & 0.692 & -- \\
      Ours (standard conv)        & 1.95M & \textbf{0.786} & \textbf{0.800} \\
      Ours (separable, deployed) & 23K   & 0.738 & 0.755 \\
      \quad same model, uint8        & 23K   & 0.740 & 0.752 \\
      \bottomrule
    \end{tabular}
  \end{minipage}
\end{table}

\subsection{Effect of dehazing on edge extraction}
\label{sec:ablation}

The premise of a two-stage design is that removing smoke improves downstream edge extraction. We tested it in two settings. The SMOKE pairs carry no edge annotations, so we measured consistency instead, comparing the edge maps of the hazy and dehazed frames against the map the same network produces on the clean frame under a dilation-based match at a fixed threshold. The hazy frame yields a consistency F-measure of 0.705 and the dehazed frame 0.801, a recovery of 0.096.

\begin{figure}[b]
  \centering
  \includegraphics[width=0.46\linewidth]{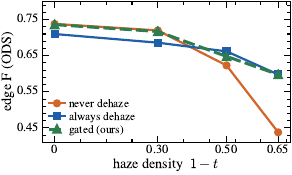}
  \caption{Edge F-measure against haze density for the three policies. Each fixed policy is the worse choice at one end of the range; the gate tracks the better of the two at each density with 43\% less arithmetic than always dehazing.}
  \label{fig:gate}
\end{figure}

The effect is not uniform across haze densities. We composited haze over the BIPEDv2 test images as $I = Jt + A(1-t)$ with $A$ fixed at 0.8, sweeping $t$ (Figure~\ref{fig:gate}). Dehazing lowers ODS on clean input (0.710 against 0.738) and under light haze (0.686 against 0.720 at $t{=}0.7$), then raises it as haze thickens, reaching 0.598 against 0.438 at $t{=}0.35$. The crossover falls between $t{=}0.7$ and $t{=}0.5$. The dehazer is a lossy reconstruction, so when haze is light enough that the edge network still recovers structure from the raw frame, the dehazer's smoothing removes more structure than the haze does.

\subsection{Gated execution}
\label{sec:gateresults}

Table~\ref{tab:gate} compares the three policies. Always dehazing is the worse choice on clear and lightly hazed frames and never dehazing is the worse choice under dense smoke. The gate tracks the better of the two at each density, staying within 0.003 ODS of never dehazing on clean and lightly hazed input, falling 0.014 ODS short of always dehazing at $t{=}0.5$, and matching it at $t{=}0.35$. It is therefore never the best policy at a single condition, but it is best of the three in the equal-weight mean. It recovers 87\% of the gap between never dehazing and a per-image oracle, and 61\% of the gap from always dehazing to that oracle. The oracle picks the better option for each image individually, while ODS scores the whole set at one shared threshold, so it is not a strict upper bound and sits 0.002 below never dehazing on clean input. It reaches that accuracy while running the dehazer on 2\% of clean frames and all of the densest, cutting mean cost from 2.81 to 1.60 GMAC per frame.

\begin{table}[h]
  \caption{Execution policies over four haze conditions, BIPEDv2 test split, one-to-one pixel matching, deployed 23K model. The gate fires on 2\%, 14\%, 72\%, and 100\% of frames respectively.}
  \label{tab:gate}
  \centering
  \footnotesize
  \begin{tabular}{lccccc}
    \toprule
    Policy & clean & $t{=}0.7$ & $t{=}0.5$ & $t{=}0.35$ & mean ODS \\
    \midrule
    Never dehaze   & \textbf{0.738} & \textbf{0.720} & 0.623 & 0.438 & 0.630 \\
    Always dehaze  & 0.710 & 0.686 & \textbf{0.662} & \textbf{0.598} & 0.664 \\
    Gated (ours)   & 0.736 & 0.717 & 0.648 & \textbf{0.598} & \textbf{0.675} \\
    \midrule
    Per-image oracle & 0.736 & 0.723 & 0.669 & 0.598 & 0.681 \\
    \bottomrule
  \end{tabular}
\end{table}

\paragraph{On-device latency.} On the Raspberry Pi (Cortex-A72, 32-bit Raspberry Pi OS, tflite-runtime 2.11, 4 threads), per-frame medians over 100 iterations are 90.6 ms for the edge network, 469.6 ms for the dehazer and 568.8 ms for both, and the gate statistic takes 10.1 ms over 200 iterations; a 120 s soak sustained 10.9 FPS on the edge stage with no thermal throttling (46 to 61\,\textdegree C). Composing the per-stage measurements with the gate's firing rates gives 110 ms on clean input and 570 ms under dense haze, a mean of 321 ms against a measured 569 ms always dehazing. The 43.5\% reduction is close to the 43.1\% predicted from arithmetic, and moves the pipeline from a uniform 1.8 FPS to 9 FPS when conditions are clear.

\section{Limitations and future work}
\label{sec:limitations}
The evaluation sets are small, consisting of only 12 real smoke pairs and 50 BIPEDv2 images. A paired bootstrap over the 50 test images places the gate's 0.011 ODS margin over always dehazing in $[0.005, 0.017]$ with 95\% confidence, while the absolute ODS of a single policy is resolved only to $\pm0.013$ at this size (Appendix~\ref{app:release}). We weight the four haze conditions equally, a modelling choice that densities recorded on real firegrounds would replace.

The latency results use the uint8 dehazer, while Table~\ref{tab:main} and the policy comparison in Table~\ref{tab:gate} use the float32 model, which the quantized version trails by 1.8 dB, enough that it falls below AOD-Net on both metrics. To recover part of that gap, we plan to implement quantization-aware training. Both the crossover and $\tau$ were calibrated on synthetic composites with constant atmospheric light, and the smoke in our captures was produced by smoke machines, so $\tau$ needs refitting on data from real fires. Testing with firefighters in fireground conditions is the last step and will require ethics review and departmental safety approval.

\section{Conclusion}
\label{sec:conclusion}

We built a two-stage vision pipeline for smoke-obscured environments that runs on a Raspberry Pi 4, and tested whether its enhancement stage always improves the downstream task. Dehazing before edge detection improves edge accuracy under dense smoke and degrades it on clear input, so the right choice varies frame to frame. A haze test costing 2\% of the dehazer's latency decides whether to run it, which is more accurate on average than either fixed choice and 43.5\% cheaper. The same approach should apply wherever an enhancement stage is expensive and only sometimes needed.

\bibliographystyle{plainnat}
\bibliography{references}

\appendix

\section{Network architectures}
\label{app:arch}

Table~\ref{tab:arch} specifies both networks. Both take $256\times256\times3$ input. In the dehazer, every convolution except the output layer is followed by LeakyReLU and batch normalization, with dropout 0.5 during training. The fourth encoder stage is a dilated convolution at rate 2 rather than a downsampling stage, so the encoder reduces resolution by $4\times$ in total and the decoder's two transposed convolutions restore it. Markers $s_1$ and $s_2$ denote skip connections: $\rightarrow$ at the stage that produces the features, $\oplus$ where the decoder concatenates them. Separable convolutions in the edge network are a $3\times3$ depthwise convolution followed by a $1\times1$ pointwise convolution, except the output layer, whose depthwise stage is $1\times1$.

The dehazer's parameter count is 353{,}251 in convolution weights and biases plus 1{,}792 in batch normalization (448 channels $\times$ 4), giving the 355{,}043 total reported in Section~\ref{sec:system}.

\begin{table}[h]
  \caption{Layer specification for both networks. Stride is 1 unless noted.}
  \label{tab:arch}
  \centering
  \footnotesize
  \begin{minipage}[t]{0.485\linewidth}
    \centering
    \textbf{Smoke removal (355{,}043 params)}\\[2pt]
    \scriptsize
    \begin{tabular}{clr}
      \toprule
      \# & Layer & Output \\
      \midrule
      1 & Conv $3\times3$, 32 \quad $\rightarrow s_1$   & $256^2\times32$ \\
      2 & Conv $3\times3$, 64, stride 2 $\rightarrow s_2$ & $128^2\times64$ \\
      3 & Conv $3\times3$, 128, stride 2               & $64^2\times128$ \\
      4 & Conv $3\times3$, 128, dilation 2             & $64^2\times128$ \\
      5 & TransposeConv $3\times3$, 64 $\oplus s_2$    & $128^2\times128$ \\
      6 & TransposeConv $3\times3$, 32 $\oplus s_1$    & $256^2\times64$ \\
      7 & Conv $3\times3$, 3, sigmoid                  & $256^2\times3$ \\
      \bottomrule
    \end{tabular}
  \end{minipage}\hfill
  \begin{minipage}[t]{0.485\linewidth}
    \centering
    \textbf{Edge detection (23{,}228 params)}\\[2pt]
    \scriptsize
    \begin{tabular}{clr}
      \toprule
      \# & Layer & Output \\
      \midrule
      1 & SepConv 32 \quad $\rightarrow s_1$      & $256^2\times32$ \\
      2 & SepConv 32, stride 2                    & $128^2\times32$ \\
      3 & SepConv 64 \quad $\rightarrow s_2$      & $128^2\times64$ \\
      4 & SepConv 64, stride 2                    & $64^2\times64$ \\
      5 & Nearest-neighbor $2\times$ $\oplus s_2$ & $128^2\times128$ \\
      6 & SepConv 64                              & $128^2\times64$ \\
      7 & Nearest-neighbor $2\times$ $\oplus s_1$ & $256^2\times96$ \\
      8 & SepConv 32                              & $256^2\times32$ \\
      9 & SepConv 1, sigmoid                      & $256^2\times1$ \\
      \bottomrule
    \end{tabular}
  \end{minipage}
\end{table}

\section{Operator support on the deployed runtime}
\label{app:opset}

Our first TensorFlow Lite exports of the edge network emitted \texttt{TILE} version 3, because \texttt{UpSampling2D} lowers to a dynamic-shape chain (\texttt{SHAPE} $\rightarrow$ \texttt{STRIDED\_SLICE} $\rightarrow$ \texttt{TILE}) rather than to a static resize. The deployment target could not register that operator and aborted inside \texttt{allocate\_tensors()}.

The limit came from the interpreter available on the device. The Raspberry Pi runs 32-bit Raspberry Pi OS Buster on Python 3.7, which pins the interpreter to \texttt{tflite-runtime} 2.11, the last release with a \texttt{cp37} \texttt{armv7l} wheel. Newer runtimes exist for this board but require a newer Python version.

We therefore rebuilt the network with an explicit nearest-neighbor resize to literal output sizes, which lowers to \texttt{RESIZE\_NEAREST\_NEIGHBOR}. Neither layer carries weights, so the transfer is exact, with a maximum absolute difference of $0.0$ between the original and rebuilt float models. The re-exported model is 53 KB, and its operator set falls from ten distinct operators to six. It scores ODS 0.741 against the original export's 0.740, a difference within requantization noise.

\section{Release, licences, and broader impact}
\label{app:release}

\paragraph{Code and data.} We will release the training and evaluation scripts, the two quantized TensorFlow Lite models, and the Raspberry Pi benchmark kit. The kit contains both uint8 models, the bundled \texttt{tflite-runtime} wheel for Python 3.7, a sample frame, and the exact commands used to produce every latency figure in Section~\ref{sec:results}. All four datasets are public and cited in Section~\ref{sec:setup}; we redistribute none of them.

\paragraph{Compute.} Every training run used a single NVIDIA GeForce RTX 3060 Ti, with each edge detection and dehazing trial completing in about five minutes, which is what made a sweep over architecture and augmentation variants practical. Evaluation ran on a workstation CPU; only the latency figures in Section~\ref{sec:results} were measured on the Raspberry Pi.

\paragraph{Resolution of the accuracy comparison.} The confidence interval quoted in Section~\ref{sec:limitations} comes from a paired bootstrap: the 50 BIPEDv2 test images are resampled with replacement 2{,}000 times, and for each resample the mean ODS of every policy is recomputed from per-image match counts, with the operating threshold re-selected each time. Because both policies see the same resampled images, the difference between them is far better resolved than either absolute value. The bootstrap measures variation across test images only, not across training seeds.

\paragraph{Dataset licences.} RESIDE~\citep{li2019reside} and SMOKE~\citep{jin2022structure} publish no formal licence and ask only for citation. DENSE-HAZE~\citep{ancuti2019dense,ancuti2019ntire} grants permission to use and publish its images and asks that both of its papers be cited. BIPEDv2~\citep{soria2023dexined} is released for non-commercial use only. Our deployed edge network is trained on BIPEDv2 and inherits that restriction, so the released models are for research use and not for commercial deployment.

\paragraph{Broader impact.} This is a research prototype. It has not been certified against any firefighting equipment standard and would need further testing and development before operational use.

\section{Output type and quantization cost}
\label{app:quant}

Quantization does not cost both stages equally due to the type of output produced. Uint8 resolution is ample because the edge map is near-binary. The dehazer regresses full RGB radiance, and its per-tensor scales must span the color range, which leaves far less resolution per level. The cost of quantization therefore has to be measured for each stage separately.

\end{document}